\documentclass[manuscript,screen]{acmart}
\usepackage{xspace}
\newcommand{\sysname}{\textsc{DeepTrans Studio}\xspace}
\usepackage{graphicx}  
\AtBeginDocument{%
  }

\setcopyright{acmlicensed}
\copyrightyear{2026}
\acmYear{2026}
\acmDOI{XXXXXXX.XXXXXXX}
\acmConference[CSCW '26]{Companion of the 2026 Computer-Supported Cooperative Work and Social Computing}{October 10--14, 2026}{Salt Lake City, UT, USA}
\acmISBN{978-1-4503-XXXX-X/2026/10}

\begin{document}

\title[DeepTrans Studio]{\sysname: Turning Expert Interventions into Shared Team Knowledge in Agentic Translation Workflows}


\author{Ziyang Lian}
\email{1257115202@shu.edu.cn}

\author{Qingya Zhang}
\email{zhangqingya@shu.edu.cn}

\author{Hao Wang}
\authornote{Corresponding author.}
\email{wang-hao@shu.edu.cn}

\author{Huiwen Xiong}
\email{hwxiong@shu.edu.cn}

\author{Qi Yang}
\email{yang-qi@shu.edu.cn}

\affiliation{%
  \institution{School of Computer Engineering and Science, Shanghai University}
  \city{Shanghai}
  \country{China}
}

\author{Lingyi Meng}
\email{lymeng@fl.ecnu.edu.cn}
\affiliation{%
  \institution{School of Foreign Languages, East China Normal University}
  \city{Shanghai}
  \country{China}
}

\author{Xiaoyi Gu}
\email{xiaoyi.gu@hotmail.com}
\affiliation{%
  \institution{School of Foreign Studies, Shanghai University}
  \city{Shanghai}
  \country{China}
}

\author{Rui Wang}
\email{wangrui12@sjtu.edu.cn}
\affiliation{%
  \institution{Department of Computer Science and Engineering, Shanghai Jiao Tong University}
  \city{Shanghai}
  \country{China}
}

\renewcommand{\shortauthors}{Lian et al.}


\begin{abstract}
Professional translation is often a team-based process: translators, reviewers, and project managers must coordinate terminology, legal force, and accountability across documents. Yet many LLM-based translation tools treat human corrections as isolated edits. Expert decisions made in one segment or by one member are rarely captured as reusable knowledge for the rest of the team. We present \sysname, a collaborative translation workspace that lets professionals intercept selected nodes in an agentic translation workflow, review evidence, revise AI outputs, and save approved decisions to a shared team memory. During the demo, attendees will role-play translators and reviewers, resolve preset terminology and legal-modal risks, and see how their decisions are propagated to downstream segments and surfaced in a teammate's workspace as reusable precedents. The demo illustrates how human interventions in AI-mediated work can become shared, traceable knowledge rather than one-off corrections. one-off corrections. Code and demo video: \url{https://github.com/hint-lab/deeptrans-studio}, \url{https://youtu.be/cNpafhHAEjg}.
\end{abstract}

\ccsdesc[500]{Human-centered computing~Interactive systems and tools}
\ccsdesc[300]{Human-centered computing~Collaborative and social computing systems and tools}
\ccsdesc[300]{Computing methodologies~Machine translation}

\keywords{Human-AI collaboration, collaborative translation, agentic workflows, team memory, accountability}
\begin{teaserfigure}
  \centering
  \includegraphics[width=1\textwidth]{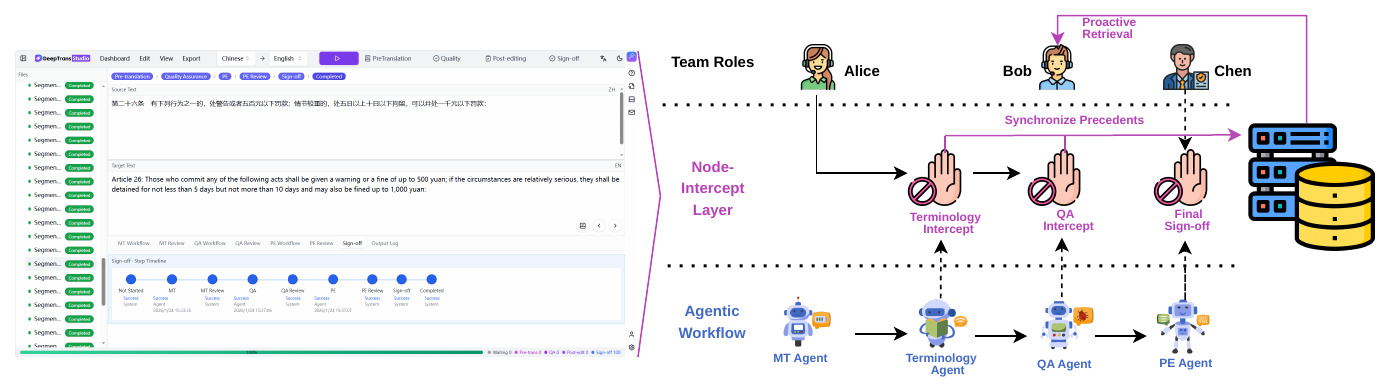} 
  \caption{Overview of DeepTrans Studio. Expert decisions made at terminology, QA, and sign-off intercepts are saved to shared team memory and later retrieved as precedents in teammate workflows. The demo lets attendees experience this flow by correcting a preset legal risk and observing its reuse in another workspace.}
  \Description{A composite image illustrating the DeepTrans Studio system. The left side displays the shared translation workspace with active workflow nodes. The right side shows a collaborative infrastructure where team roles intercept an agentic pipeline at terminology, QA, and sign-off stages. Approved expert interventions are shown synchronizing into a shared team memory, which are then proactively retrieved as reusable precedents for teammate workflows.}
  \label{fig:teaser}
\end{teaserfigure}

\maketitle

\section{Introduction}
Professional translation in high-stakes domains is collaborative work. Translators, reviewers, project managers, and clients must coordinate terminology, legal force, and accountability across people and documents. The central challenge is therefore not only to produce fluent text, but to make translation decisions consistent, visible, and reusable within a team \cite{freitag2021experts,dourish1992awareness}.

LLM-based translation tools make this coordination harder when they are designed as end-to-end pipelines or one-person chat interfaces \cite{he2024exploring, amershi2019guidelines}. Professionals may correct AI-generated drafts or multi-agent outputs \cite{briakou2024translating, wu2024transagents}, but these corrections often stay inside a local session. As a result, teams repeatedly resolve the same terminology conflicts and legal-modal ambiguities without a shared record, leading to terminological drift, duplicated review work, and accountability gaps \cite{thai2022exploring}. \sysname addresses this by making expert interventions \textbf{visible, reusable, and auditable across professional roles}.

To achieve this, the system exposes selected agentic workflow nodes for human interception, balancing automation with expert agency \cite{heer2019agency}. At terminology, QA, and sign-off points, experts can inspect evidence, revise AI suggestions, and record the rationale for their decisions\cite{bernstein2010soylent}. Approved interventions are then synchronized into a living team memory, where they guide downstream segments and future work by other members. In this way, the system shifts AI-assisted translation from isolated editing toward shared, traceable team knowledge building.

Our core contributions as a CSCW Demo include:
\begin{itemize}
    \item An interactive demonstration of \textbf{human-interceptable agentic workflows}, allowing professionals to pause selected steps and revise outputs before downstream propagation.
    \item A \textbf{shared team memory mechanism} that turns individual expert interventions into reusable precedents across segments and team members.
    \item A \textbf{role-play demo scenario} showing how translators, reviewers, and project managers coordinate AI-mediated decisions in a high-accountability workflow.
\end{itemize}

\section{System Design: Collaborative Translation Infrastructure}
To support team coordination, \sysname implements a stateful dual-loop framework that combines background automation with human-supervised decision points. The system moves beyond ``black-box'' pipelines through four collaborative modules:

\textbf{Shared Translation Workspace}: Translators, reviewers, and managers work with synchronized document states (e.g., intercepted, signed-off), establishing workflow awareness without reconstructing decisions from local chat histories.

\textbf{Node-Intercept Layer}: Instead of autonomous completion, the pipeline pauses at high-risk nodes (e.g., terminology conflicts, legal-modal ambiguities) to expose intermediate outputs for human review before downstream propagation.

\textbf{Shared Team Memory}: Expert interventions are synchronized into a ``Living Dictionary.'' These approved decisions become reusable precedents retrieved for teammate workflows, reducing repeated edits and terminological drift.

\textbf{Accountability Trace}: The system logs the provenance of model suggestions, human edits, and reused precedents. This trace allows reviewers to inspect decision chains and verify accountability across roles.





\begin{figure}[t]
  \centering
  \includegraphics[width=\columnwidth]{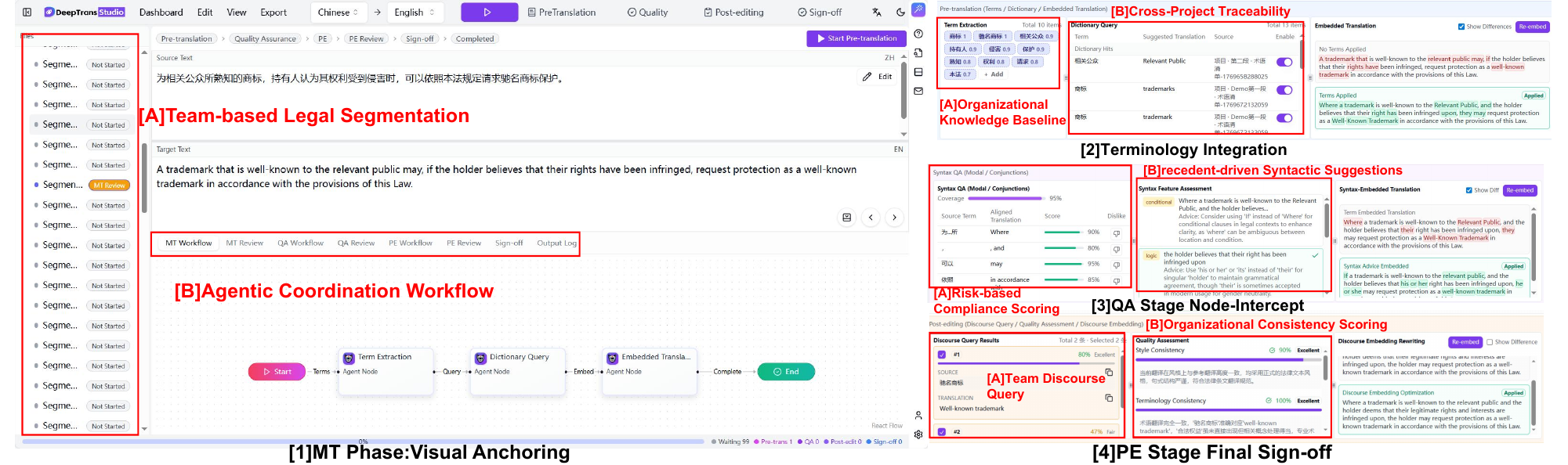}
  \Description{A four-part composite image illustrating the collaborative workflow in DeepTrans Studio. Subfigure (a) displays the shared translation workspace that anchors the source document and the multi-agent workflow. Subfigure (b) shows a terminology alignment view where an expert's resolution of conflicting evidence is synchronized into a shared team memory. Subfigure (c) illustrates a node-intercept pausing the workflow at a high-risk segment for human intervention. Subfigure (d) presents the final sign-off dashboard, allowing reviewers to inspect the accountability trace of the decision chain.}
  \caption{From individual intervention to team-wide propagation: a translator intercepts an Al decision, reviews evidence, approves a terminology choice, and the system reuses the decision in downstream segments and teammate workflows.}
  \label{fig:walkthrough}
\end{figure}

\section{\sysname in Action: A Collaborative Walkthrough}
To illustrate the collaborative workflow, we use a scenario involving a senior translator (Alice), a junior translator (Bob), and a project manager (Chen).

\textbf{1. MT Phase and Visual Anchoring}: Alice uploads a contract. The system preserves key structural cues and initializes a multi-panel workspace (Fig.~\ref{fig:walkthrough}(a)). This helps the team maintain a shared view of nested legal conditions rather than treating translation segments as isolated sentences.

\textbf{2. Terminology Intercept and Memory Sync}: Alice reviews inconsistent terminology through the alignment interface (Fig.~\ref{fig:walkthrough}(b)). After she approves a term, the decision and its rationale are synchronized into the ``Living Dictionary,'' turning a local judgment into shared team memory.

\textbf{3. Modal Intercept and Accountability}: The system triggers a \textit{node-intercept} at a legally ambiguous modal phrase, such as \textit{shall} versus \textit{may} (Fig.~\ref{fig:walkthrough}(c)). Alice inspects the surfaced evidence, revises the AI output, and records the decision before it propagates to downstream segments.

\textbf{4. Team Reuse and Final Sign-off}: When Bob later encounters a similar clause, \sysname retrieves Alice's precedent and displays it in his workspace, reducing duplicated research and aligning team decisions. Chen then uses the dashboard to inspect the \textbf{Accountability Trace} and verify the decision chain before final sign-off (Fig.~\ref{fig:walkthrough}(d)).

\section{Demo Setup and Preliminary Feedback}

\textbf{Interactive Demo}: At CSCW 2026, attendees will take part in a 5-minute role-play. First, they choose one of two preset risks in a legal contract: a terminology conflict or a legal-modal ambiguity. Next, \sysname triggers a \textit{node-intercept}, and attendees act as reviewers by approving, revising, or rejecting the AI proposal. The system then switches to a teammate view and shows how the approved intervention is retrieved as a precedent for a later segment. This setup demonstrates that attendees are not merely correcting an AI output; they are shaping how the team handles similar decisions afterward.

\textbf{Preliminary Feedback}: Formative walkthroughs with 12 professionals suggest that \sysname may help shift corrections from isolated edits to reusable team decisions. Participants especially valued the ability to inspect why a segment was interrupted, record the rationale for a correction, and make that correction visible to teammates. One senior translator (P7) noted: ``I no longer just clean machine messes; my intercepts now anchor consistency for all members, allowing individual expertise to act as a shared knowledge asset.''

\section{Conclusion and Future Work}
\sysname demonstrates how expert interventions in agentic translation workflows can be recorded as shared, traceable, and reusable team knowledge. The CSCW demo will let attendees experience how node-level intervention, team memory, and accountability tracing can support collaborative work around AI-generated outputs. Future work will extend this infrastructure to other high-stakes domains where expert judgment must be coordinated across teams.
\begin{acks}
The authors used Gemini to improve the grammar and clarity of author-written text. All system design, evaluation descriptions, and claims were written and verified by the authors, who reviewed and edited all generated suggestions and take full responsibility for the final content.
\end{acks}
This work was supported by the National Natural Science Foundation of China under Grant No. 62306173.
\bibliographystyle{ACM-Reference-Format}
\bibliography{sample-base}

@String{Computing = "Computing" }

@inproceedings{thai2022exploring,
  title={Exploring document-level literary machine translation with parallel paragraphs from world literature},
  author={Thai, Katherine and Karpinska, Marzena and Krishna, Kalpesh and Ray, Bill and Inghilleri, Moira and Wieting, John and Iyyer, Mohit},
  booktitle={Proceedings of the 2022 Conference on Empirical Methods in Natural Language Processing},
  pages={9882--9902},
  year={2022}
}

@inproceedings{wu2024transagents,
  title={TransAgents: Build your translation company with language agents},
  author={Wu, Minghao and Xu, Jiahao and Wang, Longyue},
  booktitle={Proceedings of the 2024 Conference on Empirical Methods in Natural Language Processing: System Demonstrations},
  pages={131--141},
  year={2024}
}

@inproceedings{briakou2024translating,
  title={Translating step-by-step: Decomposing the translation process for improved translation quality of long-form texts},
  author={Briakou, Eleftheria and Luo, Jiaming and Cherry, Colin and Freitag, Markus},
  booktitle={Proceedings of the Ninth Conference on Machine Translation},
  pages={1301--1317},
  year={2024}
}

@article{he2024exploring,
  title={Exploring human-like translation strategy with large language models},
  author={He, Zhiwei and Liang, Tian and Jiao, Wenxiang and Zhang, Zhuosheng and Yang, Yujiu and Wang, Rui and Tu, Zhaopeng and Shi, Shuming and Wang, Xing},
  journal={Transactions of the Association for Computational Linguistics},
  volume={12},
  pages={229--246},
  year={2024},
  publisher={MIT Press One Broadway, 12th Floor, Cambridge, Massachusetts 02142, USA~…}
}

@article{freitag2021experts,
  title={Experts, errors, and context: A large-scale study of human evaluation for machine translation},
  author={Freitag, Markus and Foster, George and Grangier, David and Ratnakar, Viresh and Tan, Qijun and Macherey, Wolfgang},
  journal={Transactions of the Association for Computational Linguistics},
  volume={9},
  pages={1460--1474},
  year={2021},
  publisher={MIT Press One Rogers Street, Cambridge, MA 02142-1209, USA journals-info~…}
}

@inproceedings{amershi2019guidelines,
  title={Guidelines for human-AI interaction},
  author={Amershi, Saleema and Weld, Dan and Vorvoreanu, Mihaela and Fourney, Adam and Nushi, Besmira and Collisson, Penny and Suh, Jina and Iqbal, Shamsi and Bennett, Paul N and Inkpen, Kori and others},
  booktitle={Proceedings of the 2019 chi conference on human factors in computing systems},
  pages={1--13},
  year={2019}
}

@article{heer2019agency,
  title={Agency plus automation: Designing artificial intelligence into interactive systems},
  author={Heer, Jeffrey},
  journal={Proceedings of the National Academy of Sciences},
  volume={116},
  number={6},
  pages={1844--1850},
  year={2019},
  publisher={National Academy of Sciences}
}

@inproceedings{dourish1992awareness,
  title={Awareness and coordination in shared workspaces},
  author={Dourish, Paul and Bellotti, Victoria},
  booktitle={Proceedings of the 1992 ACM conference on Computer-supported cooperative work},
  pages={107--114},
  year={1992}
}

@inproceedings{bernstein2010soylent,
  title={Soylent: a word processor with a crowd inside},
  author={Bernstein, Michael S and Little, Greg and Miller, Robert C and Hartmann, Bj{\"o}rn and Ackerman, Mark S and Karger, David R and Crowell, David and Panovich, Katrina},
  booktitle={Proceedings of the 23nd annual ACM symposium on User interface software and technology},
  pages={313--322},
  year={2010}
}

\end{document}